\def\paperlanguage{}
\pdfoutput=1 % for arxiv

\documentclass[letterpaper, 10 pt, conference]{ieeeconf}  % Comment this line out if you need a4paper

\usepackage{bm}
\usepackage{cite}
\include{preamble}

\newcommand{\ctext}[1]{\raise0.2ex\hbox{\textcircled{\scriptsize{#1}}}}

\IEEEoverridecommandlockouts                              % This command is only needed if
\title{\LARGE \textbf
  {
    \switchlanguage%
    {%
      Design Optimization of Three-Dimensional Wire Arrangement Considering Wire Crossings for Tendon-driven Robots
    }%
    {%
      腱駆動ロボットにおけるワイヤ交差を考慮した3次元ワイヤ配置最適化
    }%
  }
}

\author{Kento Kawaharazuka$^{1}$, Shintaro Inoue$^{1}$, Yuta Sahara$^{1}$, Keita Yoneda$^{1}$, Temma Suzuki$^{1}$, and Kei Okada$^{1}$% <-this % stops a space
  \thanks{$^{1}$ The authors are with the Department of Mechano-Informatics, Graduate School of Information Science and Technology, The University of Tokyo, 7-3-1 Hongo, Bunkyo-ku, Tokyo, 113-8656, Japan.
    {\texttt\small [kawaharazuka, s-inoue, sahara, yoneda, t-suzuki, k-okada]@jsk.imi.i.u-tokyo.ac.jp}
  }
}
\begin{document}

\maketitle
\thispagestyle{empty}
\pagestyle{empty}

%%%%%%%%%%%%%%%%%%%%%%%%%%%%%%%%%%%%%%%%%%%%%%%%%%%%%%%%%%%%%%%%%%%%%%%%%%%%%%%%
\begin{abstract}
  \switchlanguage%
  {%
    Tendon-driven mechanisms are useful from the perspectives of variable stiffness, redundant actuation, and lightweight design, and they are widely used, particularly in hands, wrists, and waists of robots.
    The design of these wire arrangements has traditionally been done empirically, but it becomes extremely challenging when dealing with complex structures.
    Various studies have attempted to optimize wire arrangement, but many of them have oversimplified the problem by imposing conditions such as restricting movements to a 2D plane, keeping the moment arm constant, or neglecting wire crossings.
    Therefore, this study proposes a three-dimensional wire arrangement optimization that takes wire crossings into account.
    We explore wire arrangements through a multi-objective black-box optimization method that ensures wires do not cross while providing sufficient joint torque along a defined target trajectory.
    For a 3D link structure, we optimize the wire arrangement under various conditions, demonstrate its effectiveness, and discuss the obtained design solutions.
  }%
  {%
    腱駆動型ロボットは可変剛性や冗長駆動, 軽量化の観点から有用であり, 特にハンドや手首, 腰などに広く活用されている.
    これらのワイヤ配置の設計は人手により行われてきたが, それらは複雑な構造になると困難を極める.
    これまでワイヤ配置を自動化する様々な研究が行われてきたが, その多くは平面上の動作やモーメントアーム一定の条件が課されていたり, ワイヤの交差を考慮していなかったりと, その問題設定を単純化してきた.
    そこで本研究では, ワイヤの交差を考慮した3次元のワイヤ配置最適化を提案する.
    設定した目標軌道においてワイヤが交差せず, かつ十分な関節トルクを発揮可能なワイヤ配置をブラックボックスな多目的最適化により探索する.
    3次元リンク構造に対して, 多様な条件下でワイヤ配置を最適化し, その有効性を示すとともに, 得られた設計解について考察する.
  }%
\end{abstract}

\section{INTRODUCTION}\label{sec:introduction}
\switchlanguage%
{%
  Various tendon-driven robots have been developed to date.
  These range from biomimetic robots that closely mimic the human skeleton and muscles \cite{jantsch2013anthrob, asano2016kengoro, kawaharazuka2019musashi} to non-biomimetic robots that effectively leverage the advantages of wires \cite{endo2019superdragon, kim2015lims, temma2024saqiel, sinoue2024cubix}.
  Tendon-driven mechanisms offer several useful characteristics, such as redundant actuation enabling continued operation even when one wire breaks, variable stiffness control through hardware, and lightweight design of the distal end by allowing all actuators to be arranged in the root link.
  These features are being effectively utilized, especially in areas such as hands, wrists, and waists.
  However, the design, particularly the wire arrangement, is still mostly done empirically, and designing becomes extremely difficult as the structure becomes more complex.

  In response to this, various attempts have been made to optimize wire arrangement.
  These wire arrangement optimization methods can be broadly categorized into two conditions: constant moment arm and variable moment arm.
  The constant moment arm refers to systems in which pulleys are placed at the joints, and the distance from the wire to the joint center remains constant.
  In the constant moment arm condition, the optimization mainly focuses on the radius and spacing of the pulleys, which correspond to the muscle Jacobian.
  In \cite{pollard2002arrangement}, the muscle Jacobian is numerically optimized to secure a torque space equivalent to that of a human hand for robotic fingers.
  In \cite{asaoka2012arrangement}, the presence or absence of moment arms for each muscle is optimized for a musculoskeletal robot with an exhaustive search.
  In \cite{roozing2016arrangement}, the pulley radius, elasticity, and pre-tension are optimized for the design of asymmetric compliance actuators.
  In \cite{dong2018arrangement}, the spacing between the two fingers, pulley radius, and pulley spacing are optimized for a two-fingered hand by a genetic algorithm.
  In \cite{kawaharazuka2021redundancy} the muscle Jacobian is optimized for a musculoskeletal robot using a genetic algorithm, ensuring continued operation even if one muscle is severed.
  In \cite{islam2024tendon}, the pulley radius, link length, and pre-tension of the tendon-driven underactuated robot are optimized using reinforcement learning.

  In the variable moment arm condition, the positions of the start point, relay points, and end points of the wires are optimized.
  In \cite{rayne2018arrangement}, the distance between joints and wires is optimized to enlarge the feasible joint angle space for continuum robots.
  In \cite{hamida2021arrangement}, the wire attachment points of a Cable-Driven Parallel Robot (CDPR) is optimized for upper limb rehabilitation based on evolutionary computation and multiple objective functions.
  In \cite{jamshidifar2017arrangement}, the tension and wire attachment points are numerically optimized for the stiffness adjustment of CDPR.
  In \cite{zhong2022arrangement}, the positions of start points and end points of wires are optimized for a musculoskeletal robot to construct a constraint force field.
  In \cite{kawaharazuka2024arrangeopt}, a complex wire arrangement optimization method that includes not only start and end points but also relay points of wires is proposed.
}%
{%
  これまで様々な腱駆動型ロボットが開発されてきた.
  それらは人体を模倣した筋骨格ロボット\cite{jantsch2013anthrob, asano2016kengoro, kawaharazuka2019musashi}から, ワイヤの利点を上手く活かしたワイヤ駆動ロボット\cite{sinoue2024cubix, kim2015lims, temma2024saqiel}まで様々である.
  ワイヤには, 冗長駆動によるワイヤ破断時の継続的動作やハードウェアによる可変剛性, ワイヤを全てルートリンクに配置して先端を軽量化できるなどの特性があり, これらが有効に活用されている.
  特にハンドや手首, 腰回りなどには頻繁に腱駆動機構が採用されている.
  一方で, 基本的にその設計, 特にワイヤ配置は人間の手作業に委ねられている場合がほとんどであり, 複雑な構造になるとその設計は困難を極める.

  これに対して, ワイヤ配置を自動化しようという様々な試みがなされている.
  これらワイヤ配置最適化の手法は主にモーメントアーム一定, またはモーメントアーム可変の2種類の条件に大別される.
  モーメントアーム一定とは関節にプーリが配置され, ワイヤの関節中心に対する距離が一定な系, モーメントアーム可変はそうではない系である.
  モーメントアームが一定の条件化では, 主にプーリの半径やプーリ間隔, つまり筋長ヤコビアンが最適化される.
  \cite{pollard2002arrangement}はロボットフィンガについて, 人間と同等なトルク空間を確保するために, 筋長ヤコビアンを数値的に最適化している.
  \cite{asaoka2012arrangement}は筋骨格型のロボットについて, 各筋の各関節に対するモーメントアームの有無を全探索で最適化している.
  \cite{roozing2016arrangement}は非対称コンプライアンスアクチュエータの設計に向けてワイヤのプーリ半径や弾性, プリテンションを数値的に最適化している.
  \cite{dong2018arrangement}は二指ハンドの間隔やプーリ半径 プーリ間隔を含む腱配置の最適化を遺伝的アルゴリズムに基づき行っている.
  \cite{kawaharazuka2021redundancy}は筋骨格型のロボットについて, どの筋が切れても動作継続可能な筋長ヤコビアンを遺伝的アルゴリズムにより最適化している.
  \cite{islam2024tendon}は劣駆動型の腱駆動ロボットについて, プーリ半径とリンク長さ, プリテンションを強化学習により最適化している.

  これに対して, モーメントアーム可変の場合はワイヤの始点や中継点, 終止点が最適化される.
  \cite{rayne2018arrangement}はcontinuumロボットについて, feasible joint angleの空間をenlargeするために, 関節とワイヤの距離を探索的に最適化している.
  \cite{hamida2021arrangement}は上肢のリハビリテーションを行うためのCable-Driven Parallel Robot (CDPR)について, 複数の目的関数からそのワイヤ取り付け位置を進化計算に基づき最適化している.
  \cite{jamshidifar2017arrangement}はCDPRについて, その剛性最適化に向けた張力やワイヤ取り付け位置を数値的に最適化している.
  \cite{zhong2022arrangement}は筋骨格型のロボットについて, Constraint Force Field構築に向けて筋の起始点と終止点の位置を数値的に最適化している.
  \cite{kawaharazuka2024arrangeopt}は筋の起始点と終止点 だけでなく, 中継点まで含めた複雑なワイヤ配置最適化手法を提案している.
}%

\begin{figure}[t]
  \centering
  \includegraphics[width=0.8\columnwidth]{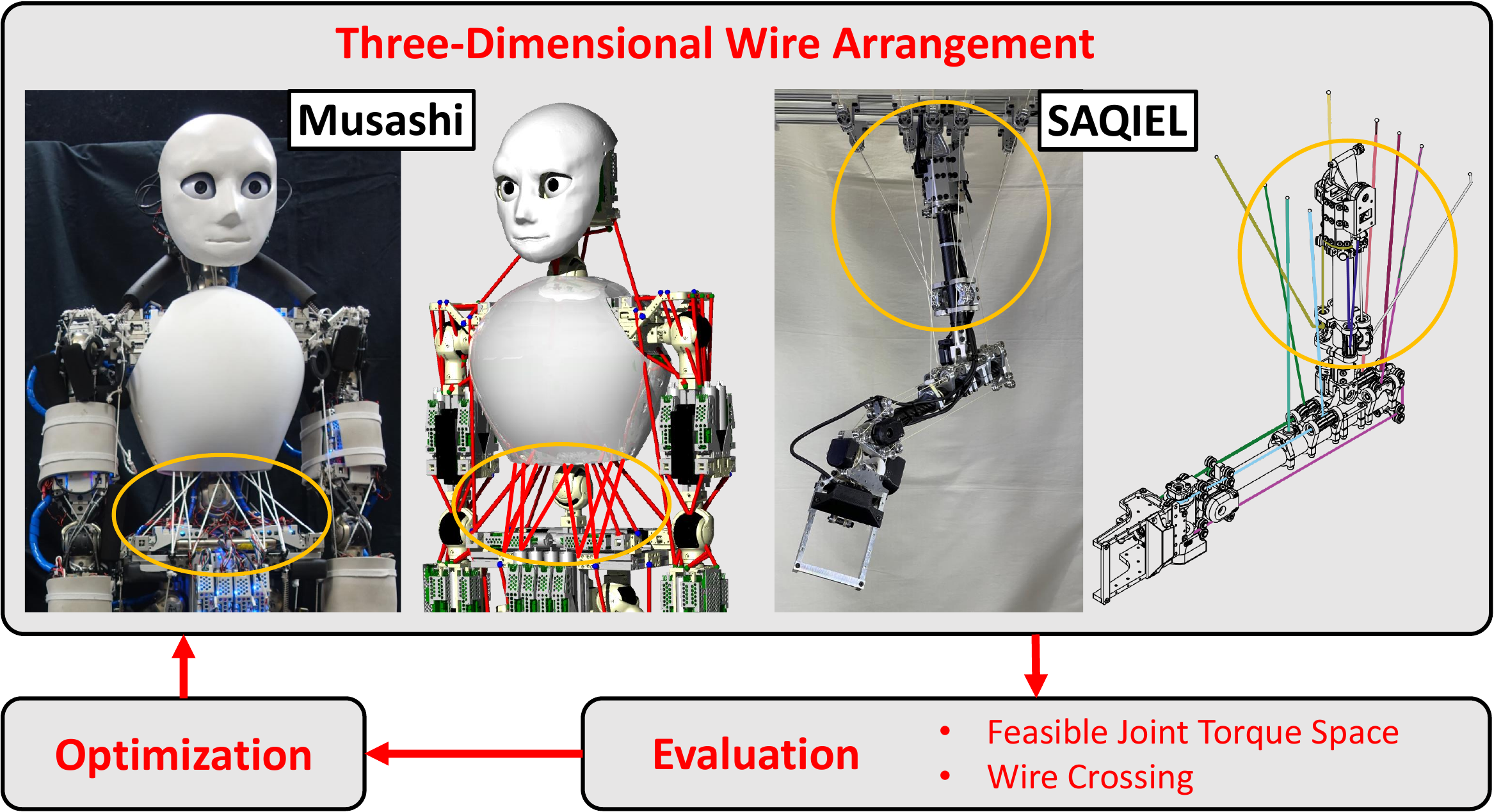}
  \vspace{-1.0ex}
  \caption{The concept of this study: complex three-dimensional wire arrangement can be optimized with multi-objective optimization considering the wire crossings and feasible joint torque space. The figures show the waist of tendon-driven musculoskeletal humanoid Musashi \cite{kawaharazuka2019musashi} and the shoulder of light-weight safe manipulator SAQIEL \cite{temma2024saqiel} that we have developed so far.}
  \vspace{-4.0ex}
  \label{figure:concept}
\end{figure}

\begin{figure*}[t]
  \centering
  \includegraphics[width=1.7\columnwidth]{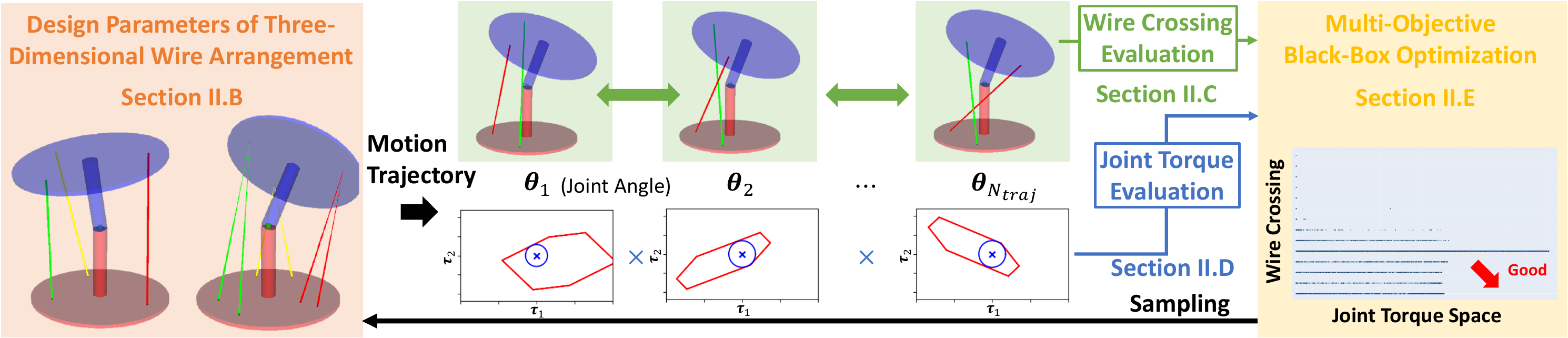}
  \vspace{-1.0ex}
  \caption{The overview of the three-dimensional wire arrangement optimization considering wire crossings and feasible joint torque space.}
  \vspace{-3.0ex}
  \label{figure:overview}
\end{figure*}

\switchlanguage%
{%
  On the other hand, many of the previously introduced studies focus on cases where the moment arm is constant \cite{pollard2002arrangement, roozing2016arrangement, dong2018arrangement, asaoka2012arrangement, kawaharazuka2021redundancy, islam2024tendon} or a planar motion is assumed \cite{rayne2018arrangement, zhong2022arrangement, kawaharazuka2024arrangeopt}.
  While wire arrangement optimization in three-dimensional spaces with variable moment arms has been conducted for CDPR \cite{rayne2018arrangement, hamida2021arrangement, jamshidifar2017arrangement}, these studies do not consider wire crossings.
  They fine-tune the wire positions under the assumption that wire crossings do not occur in a large space.
  In other words, many methods have simplified the problem of wire arrangement optimization.
  However, in reality, there are numerous complex structures where the system is three-dimensional, the moment arms are variable, and wire crossings do occur \cite{jantsch2013anthrob, asano2016kengoro, kawaharazuka2019musashi, temma2024saqiel}.
  For example, the shoulder of SAQIEL \cite{temma2024saqiel} and the waist of Musashi \cite{kawaharazuka2019musashi} shown in \figref{figure:concept} face significant issues with wire crossings during rotation in the yaw direction.
  Similarly, this issue occurs in the freely configurable wire arrangement of the mobile tendon-driven robot CubiX \cite{sinoue2024cubix}, as well as in the shoulder, waist, and hip joints of most other musculoskeletal humanoids \cite{jantsch2013anthrob, asano2016kengoro}.
  When wires cross, not only can they become tangled and destabilize the structure, but the friction may also increase the risk of wire breakage.
  It is necessary to explore wire arrangements that prevent multiple wires from crossing within a small space, while still allowing for controllable and sufficient joint torque.

  In this study, we propose a three-dimensional wire arrangement optimization method that takes wire crossings into account (\figref{figure:concept}).
  We define a target trajectory and perform multi-objective black-box optimization to explore wire arrangements that prevent wire crossings along the trajectory while ensuring a stable joint torque space.
  We optimize wire arrangements under various conditions for a three-dimensional link structure, demonstrating the effectiveness of the method and discussing the resulting design solutions.
}%
{%
  一方で, これまで紹介した論文は, モーメントアーム一定の条件下\cite{pollard2002arrangement, roozing2016arrangement, dong2018arrangement, asaoka2012arrangement, kawaharazuka2021redundancy}や平面上の動作を前提とした研究\cite{rayne2018arrangement, zhong2022arrangement, kawaharazuka2024arrangeopt}が多くを占める.
  モーメントアームが可変かつ3次元のワイヤ配置設計最適化はCDPR型のロボットにおいて行われているが\cite{rayne2018arrangement, hamida2021arrangement, jamshidifar2017arrangement}, それらはワイヤの交差を考慮していない.
  あくまで広い空間で, ワイヤ交差が起きないことを前提に細かなワイヤ位置の調整を行っている.
  つまり, これまで様々な方法でワイヤ配置最適化に関する問題を簡単化してきたのである.
  しかし, 現実には3次元構造かつモーメントアーム可変で, ワイヤ交差が起こるような複雑な構造が多数存在している\cite{jantsch2013anthrob, asano2016kengoro, kawaharazuka2019musashi, kim2015lims, temma2024saqiel}.
  例えば\figref{figure:concept}にあるSAQIEL \cite{temma2024saqiel}の根本やMusashi \cite{kawaharazuka2019musashi}の腰はYaw方向への回転時にワイヤ同士の干渉が大きく問題になる.
  同様に, ワイヤを自由に配置可能なmobile tendon-driven robot CubiX \cite{sinoue2024cubix}や, その他ほとんどの筋骨格ヒューマノイド\cite{jantsch2013anthrob, asano2016kengoro}の肩や腰, 股関節について同様の問題が発生する.
  ワイヤが交差するとそれらは絡まり合い構造が不安定になるだけでなく, 摩擦によりワイヤ破断の危険性がある.
  そのため, 例えばロボットの手首を腱駆動構造で構築しようとした場合, 小さな空間を張る複数のワイヤが交差してしまうことを防ぎつつ, かつ可制御で十分な関節トルクを発揮可能なワイヤ配置を探索する必要がある.

  そこで本研究では, ワイヤの交差まで考慮した3次元のワイヤ配置設計最適化手法を提案する(\figref{figure:concept}).
  動作軌道を定義し, その軌道上でワイヤ同士が交差してしまうことを防ぎつつ, 安定した発揮可能関節トルク空間を確保できるよう, ワイヤ配置の多目的ブラックボックス最適化を行う.
  3次元リンク構造に対して, 多様な条件下でワイヤ配置を最適化し, その有効性を示すとともに, 得られた設計解について考察する.

  本研究の構成は以下である.
  \secref{sec:proposed}では, ワイヤ配置の設計パラメータ, ワイヤ交差の検知, 発揮可能関節トルクの計算, 多目的最適化について順に述べる.
  \secref{sec:experiment}では, 3次元リンク構造について, 多様なワイヤ配置を構築する.
  \secref{sec:discussion}では実験結果について考察し, \secref{sec:conclusion}で結論を述べる.
}%

\section{Design Optimization of Three-dimensional Wire Arrangement Considering Wire Crossings} \label{sec:proposed}
\subsection{Overview of Wire Arrangement Optimization} \label{subsec:overview}
\switchlanguage%
{%
  As in \figref{figure:overview}, we describe the design parameters of the wire arrangement in \secref{subsec:design-params}, the method for detecting wire crossings in \secref{subsec:wire-crossing}, the calculation method of feasible joint torque in \secref{subsec:torque-eval}, and the multi-objective optimization method in \secref{subsec:optimization}.
  We briefly outline the problem setting addressed in this study.
  The body structure considered is a two-link three-dimensional structure, as shown on the left side of \figref{figure:overview}.
  By placing the start, relay, and end points of the wires on each link, this structure is actuated.
  The task is to achieve a given joint angle trajectory.
  Therefore, a sequence of joint angles $\{\bm{\theta}_1, \bm{\theta}_2, \cdots, \bm{\theta}_{N_{traj}}\}$ is provided.
  $N_{traj}$ represents the number of joint angle sequences, and $\bm{\theta}_{i}$ is a $D$-dimensional vector, where $D$ is the number of DOFs of the joint.
  Here, it is required that no wire crossings occur along the trajectory and that sufficient feasible joint torque is ensured.
  The goal of this study is to explore a wire arrangement that satisfies these conditions.
}%
{%
  本研究の全体像を\figref{figure:overview}に示す.
  \secref{subsec:design-params}では, ワイヤ配置の設計パラメータについて, \secref{subsec:wire-crossing}ではワイヤ交差の検知方法について, \secref{subsec:torque-eval}では発揮可能関節トルクの計算方法について, \secref{subsec:optimization}では多目的最適化手法について述べる.

  ここで, 本研究で扱う問題設定について簡潔に述べておく.
  まず, 扱う身体構造は2リンクの3次元構造であり, \figref{figure:overview}の左に示すような構造である.
  各リンクにワイヤの始点・中継点・終止点を配置することで, この構造を駆動する.
  行うタスクは与えられた関節角度軌道を実現することである.
  そのため, 関節角度列$\{\bm{\theta}_1, \bm{\theta}_2, \cdots, \bm{\theta}_{N_{traj}}\}$ ($N_{traj}$は関節角度列の数を表す)が与えられる.
  この際, 軌道上でワイヤが交差しないこと, 十分な発揮可能関節トルクを確保することが求められる.
  これらの条件を満たすワイヤ配置を探索するのが本研究の目的である.
}%

\begin{figure}[t]
  \centering
  \includegraphics[width=0.8\columnwidth]{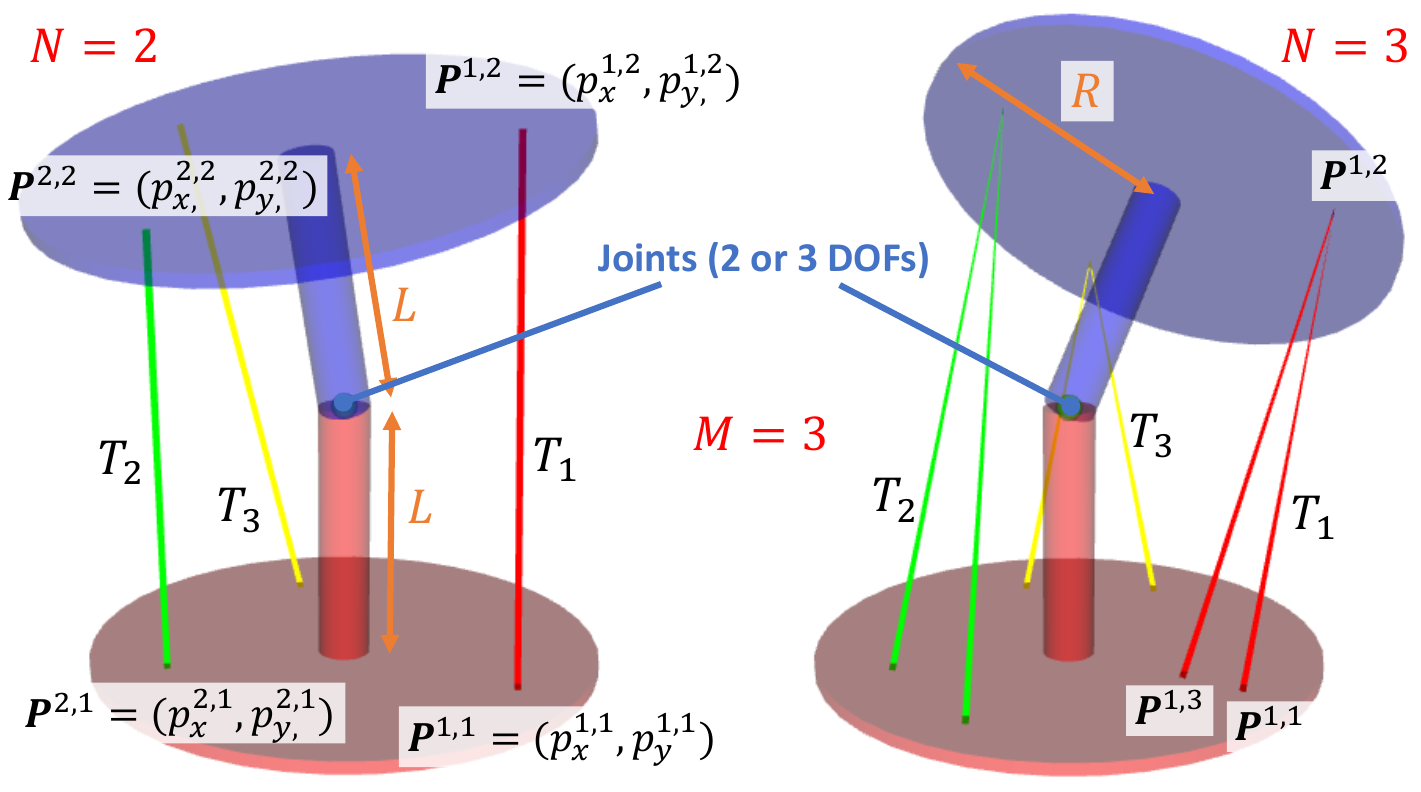}
  \vspace{-1.0ex}
  \caption{The design parameters of wire arrangement.}
  \vspace{-4.0ex}
  \label{figure:configuration}
\end{figure}

\subsection{Design Parameters of Wire Arrangement} \label{subsec:design-params}
\switchlanguage%
{%
  The detailed link structure addressed in this study and an example of wire arrangement are shown in \figref{figure:configuration}.
  This structure consists of two primary links with joints that have 2-3 degrees of freedom (DOFs), and it can be seen as a direct extension of the general two-dimensional link structures of tendon-driven robots into three dimensions.
  Each link is divided into a long cylindrical section with length $L$ and a larger circular section with radius $R$, where the start, relay, and end points of the wires can be placed.
  We arrange $M$ wires on the structure, with each wire having $N$ start, relay, and end points.
  \figref{figure:configuration} shows an example of wire arrangement with $M=3$ and $N=\{2, 3\}$.
  The position of the $j$-th relay point on the $i$-th wire is represented as $\bm{P}^{i,j}=(p^{i, j}_{x}, p^{i, j}_{y})$.
  When $N=2$, each wire connects the links in a straight line, while for $N=3$, a foldback occurs, and the start and end points are placed on the same link.
  Thus, the design parameters for the wire arrangement in this study consist of $2 \times N \times M$ continuous values $\{p^{i, j}_{x}, p^{i, j}_{y}\}$, where $1 \leq i \leq M$ and $1 \leq j \leq N$.
  The range of $p^{i, j}_{\{x, y\}}$ is constrained to the surface of the larger circular section as follows.
  \begin{align}
    -R \leq &p^{i, j}_{x} \leq R\\
    -\sqrt{R^2 - (p^{i, j}_{x})^2} \leq &p^{i, j}_{y} \leq \sqrt{R^2 - (p^{i, j}_{x})^2}
  \end{align}
}%
{%
  本研究で扱う腱駆動ロボットのワイヤ配置に関する設計パラメータについて述べる.
  詳細なリンク構造とワイヤ配置例を\figref{figure:configuration}に示す.
  これは2つの主要なリンクの間に2-3の自由度の関節を持つ構造であり, 腱駆動ロボットにおける2次元平面上の一般的なリンク構造を3次元に直接拡張した形と言える.
  各リンクは長さ$L$の細長い円筒と半径$R$の大円部に分かれており, この大円部にワイヤの始点・中継点・終止点を配置することができる.
  本研究では対象となる構造に$M$本のワイヤを配置するが, それぞれのワイヤの始点・中継点・終止点の数を$N$とする.
  \figref{figure:configuration}には, $M=3$かつ$N=\{2, 3\}$のときのワイヤ配置例が示されている.
  $i$番目のワイヤにおける$j$番目のワイヤ端点の位置を$\bm{P}^{i,j}=(p^{i, j}_{x}, p^{i, j}_{y})$とする.
  $N=2$の場合は各ワイヤがリンクを直線的に繋ぐが, $N=3$の場合は折返しが発生し, 始点と終止点が同じリンク上に配置される.
  つまり本研究におけるワイヤ配置の設計パラメータは$\{p^{i, j}_{x}, p^{i, j}_{y}\}$ ($1 \leq i \leq M, 1 \leq j \leq N$)の$2\times N \times M$個の連続値となる.
  ここで, $p^{i, j}_{\{x, y\}}$の範囲は以下のように大円上に制約される.
  \begin{align}
    -R \leq &p^{i, j}_{x} \leq R\\
    -\sqrt{R^2 - (p^{i, j}_{x})^2} \leq &p^{i, j}_{y} \leq \sqrt{R^2 - (p^{i, j}_{x})^2}
  \end{align}
}%

\subsection{Detection of Wire Crossings} \label{subsec:wire-crossing}
\begin{figure}[t]
  \centering
  \includegraphics[width=0.9\columnwidth]{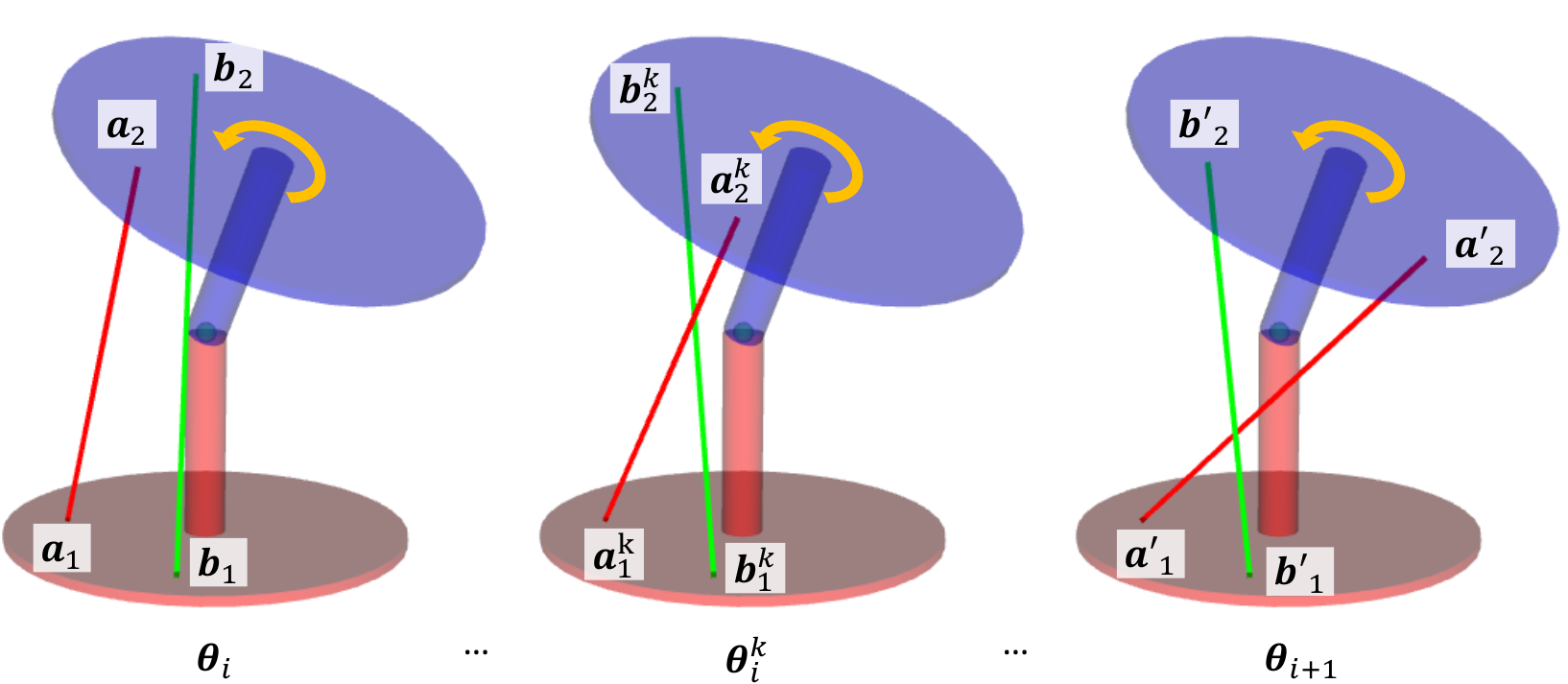}
  \vspace{-1.0ex}
  \caption{The configuration detecting wire crossings.}
  \vspace{-3.0ex}
  \label{figure:crossing}
\end{figure}

\switchlanguage%
{%
  First, we consider the minimum distance between two wires.
  Let the start point of one wire at joint angle $\bm{\theta}_{i}$ be $\bm{a}_{1}$ and the end point be $\bm{a}_{2}$, and the start point of the other wire be $\bm{b}_{1}$ and the end point be $\bm{b}_{2}$.
  For simplicity, we do not use the notation $\bm{P}^{i, j}$ here.
  Each wire segment and the distance between the two wires can be represented using parameters $s$ and $t$ ($0 \leq s, t \leq 1$) as follows.
  \begin{align}
    \bm{p}_{a}(s) &= \bm{a}_{1} + s\cdot(\bm{a}_{2}-\bm{a}_{1})\\
    \bm{p}_{b}(t) &= \bm{b}_{1} + t\cdot(\bm{b}_{2}-\bm{b}_{1})\\
    d(s, t) &= ||\bm{p}_{a}(s) - \bm{p}_{b}(t)||_{2}\nonumber\\
            &= ||\bm{a}_{1} + s\cdot\bm{u} - (\bm{b}_{1} + t\cdot\bm{v})||_{2}\nonumber\\
            &= ||\bm{w} + s\cdot\bm{u} - t\cdot\bm{v}||_{2}
  \end{align}
  where $\bm{u} = \bm{a}_{2} - \bm{a}_{1}$, $\bm{v} = \bm{b}_{2} - \bm{b}_{1}$, and $\bm{w} = \bm{a}_{1} - \bm{b}_{1}$, and $||\cdot||_{2}$ denotes the L2 norm.
  To find the values of $s$ and $t$ that minimize $d(s, t)$, we differentiate $d(s, t)$ with respect to $s$ and $t$, and solve for the points where these partial derivatives are zero.
  \begin{align}
    \frac{\partial d(s, t)}{\partial s} &= 2\bm{u}\cdot(\bm{w} + s\cdot\bm{u} - t\cdot\bm{v}) = 0\\
    \frac{\partial d(s, t)}{\partial t} &= -2\bm{v}\cdot(\bm{w} + s\cdot\bm{u} - t\cdot\bm{v}) = 0
  \end{align}
  By solving this system of equations, we can compute the values of $s_c$ and $t_c$ that minimize $d(s, t)$ as follows,
  \begin{align}
    s_c = \frac{be-cd}{ac-b^2}, t_c = \frac{ae-bd}{ac-b^2}
  \end{align}
  where $a = \bm{u}\cdot\bm{u}$, $b = \bm{u}\cdot\bm{v}$, $c = \bm{v}\cdot\bm{v}$, $d = \bm{u}\cdot\bm{w}$, and $e = \bm{v}\cdot\bm{w}$.
  Finally, considering the constraints on $s$ and $t$, the minimum distance $s^{*}_{c}$ and $t^{*}_{c}$ are calculated as follows.
  \begin{align}
    s^{*}_{c} &= \max(0, \min(1, s_c))\\
    t^{*}_{c} &= \max(0, \min(1, t_c))
  \end{align}
  The minimum distance between the two wires is then given by $d^{*} = d(s^{*}_{c}, t^{*}_{c})$.

  Next, we consider the trajectory from $\bm{\theta}_{i}$ to $\bm{\theta}_{i+1}$, and determine whether the wires cross during this movement.
  To determine if two wires cross or touch, we check whether there is any point along the trajectory where $d^{*} < \epsilon$ (where $\epsilon$ is a sufficiently small constant, set to $1.0 \times 10^{-4}$ in this study).
  The start and end points of the two wires at a certain point on the trajectory are computed as follows,
  \begin{align}
    &\bm{a}^{k}_{1} = k\bm{a}_{1} + (1-k)\bm{a}'_{1} &\bm{a}^{k}_{2} = k\bm{a}_{2} + (1-k)\bm{a}'_{2}\nonumber\\
    &\bm{b}^{k}_{1} = k\bm{b}_{1} + (1-k)\bm{b}'_{1} &\bm{b}^{k}_{2} = k\bm{b}_{2} + (1-k)\bm{b}'_{2}\nonumber
  \end{align}
  where $k$ is a continuous parameter ($0 \leq k \leq 1$), and $\{\bm{a}, \bm{b}\}'_{\{1, 2\}}$ represent the start and end points of the wires at $\bm{\theta}_{i+1}$.
  Ideally, we should compute $\{\bm{a}, \bm{b}\}^{k}_{\{1, 2\}}$ for each point along the trajectory from $\bm{\theta}_{i}$ to $\bm{\theta}_{i+1}$, but for computational efficiency, we approximate them using linear interpolation between $\{\bm{a}, \bm{b}\}_{\{1, 2\}}$ and $\{\bm{a}, \bm{b}\}'_{\{1, 2\}}$.
  By dividing the parameter $k$ and checking if $d^{*} < \epsilon$ at each step, we can detect wire crossings.
  However, setting $\epsilon$ appropriately is challenging; improper settings may fail to detect wire crossings.
  Similarly, setting the number of divisions for $k$ is also difficult; if the divisions are too coarse, there may be a failure to detect wire crossings, while if they are too fine, the computational cost becomes enormous.
  To avoid manually setting $\epsilon$ and the number of divisions for $k$, we utilize the fact that $d^{*}$ is convex downwards and employ Brent's method \cite{brent2013algorithms}, a root-finding algorithm, to search for $k$ where $d^{*} < \epsilon$.
  If the maximum number of iterations (set to 20 in this study) is exceeded without converging to a solution where $d^{*} < \epsilon$, we conclude that no wire crossing occurs.

  So far, we have discussed wire crossings between two wires.
  In practice, this is done for all possible combinations of wires.
  If there is a foldback, we also check whether different segments of the same wire cross.
  Additionally, to ensure that the wires do not cross the joints, we perform similar calculations to detect intersections between the wire segments and the joint link segments.
}%
{%
  本ロボットが与えられた関節角度列$\{\bm{\theta}_1, \bm{\theta}_2, \cdots, \bm{\theta}_{N_{traj}}\}$を動作する間に, 配置されたワイヤ同士が干渉してしまうかどうかを計算する.
  \figref{figure:crossing}のような状況を考える.
  ここでは2本のワイヤのみに着目する.

  まず, 2本のワイヤの最小距離について考える.
  $\bm{\theta}_{i}$のときの一方のワイヤの始点を$\bm{a}_{1}$, 終止点を$\bm{a}_{2}$, もう一方のワイヤの始点を$\bm{b}_{1}$, 終止点を$\bm{b}_{2}$とする(簡単のため$\bm{P}^{i, j}$という表記は使わない).
  ここで, 各ワイヤの線分はパラメータ$s$と$t$を用いて以下のように表現できる.
  \begin{align}
    \bm{p}_{a}(s) &= \bm{a}_{1} + s\cdot(\bm{a}_{2}-\bm{a}_{1})\\
    \bm{p}_{b}(t) &= \bm{b}_{1} + t\cdot(\bm{b}_{2}-\bm{b}_{1})\\
    & 0 \leq s, t \leq 1
  \end{align}
  ここで, 2本のワイヤの距離は以下のように表現できる.
  \begin{align}
    d(s, t) &= ||\bm{p}_{a}(s) - \bm{p}_{b}(t)||_{2}\nonumber\\
            &= ||\bm{a}_{1} + s\cdot\bm{u} - (\bm{b}_{1} + t\cdot\bm{v})||_{2}\nonumber\\
            &= ||\bm{w} + s\cdot\bm{u} - t\cdot\bm{v}||_{2}
  \end{align}
  ここで, $\bm{u} = \bm{a}_{2} - \bm{a}_{1}$, $\bm{v} = \bm{b}_{2} - \bm{b}_{1}$, $\bm{w} = \bm{a}_{1} - \bm{b}_{1}$, $||\cdot||_{2}$はL2ノルムを表す.
  このとき, $d(s, t)$が最小となる$s$と$t$を計算する.
  これは, $d(s, t)$を$s$と$t$で偏微分し, それぞれが0となる点を求めることで解くことができる.
  \begin{align}
    \frac{\partial d(s, t)}{\partial s} &= 2\bm{u}\cdot(\bm{w} + s\cdot\bm{u} - t\cdot\bm{v}) = 0\\
    \frac{\partial d(s, t)}{\partial t} &= -2\bm{v}\cdot(\bm{w} + s\cdot\bm{u} - t\cdot\bm{v}) = 0
  \end{align}
  この連立方程式を解くことで, 以下のように$d(s, t)$が最小となる$s_c$と$t_c$が計算できる.
  \begin{align}
    s_c = \frac{be-cd}{ac-b^2}, t_c = \frac{ae-bd}{ac-b^2}
  \end{align}
  ここで, $a = \bm{u}\cdot\bm{u}$, $b = \bm{u}\cdot\bm{v}$, $c = \bm{v}\cdot\bm{v}$, $d = \bm{u}\cdot\bm{w}$, $e = \bm{v}\cdot\bm{w}$である.
  最後に, $s$と$t$の制約条件を考慮し, 以下のように最小距離における$s^{*}_{c}$, $t^{*}_{c}$を計算する.
  \begin{align}
    s^{*}_{c} &= \max(0, \min(1, s_c))\\
    t^{*}_{c} &= \max(0, \min(1, t_c))
  \end{align}
  このときの$d^{*} = d(s^{*}_{c}, t^{*}_{c})$がワイヤ同士の最小距離となる.

  次に, ある関節角度$\bm{\theta}_{i}$から$\bm{\theta}_{i+1}$までの軌道を考え, この間にワイヤ同士が交差するかを判定する.
  ここで, 2本のワイヤが交差, または接触しているかどうかは, この軌道上で$d^{*} < \epsilon$ ($\epsilon$は十分に小さな定数, 本研究では$1.0\times10^{-4}$とした)となる瞬間があるかどうかで判定できる.
  軌道上のある点における2本のワイヤの始点・終止点を以下のように計算する.
  \begin{align}
    \bm{a}^{k}_{1} &= k\bm{a}_{1} + (1-k)\bm{a}'_{1}\\
    \bm{a}^{k}_{2} &= k\bm{a}_{2} + (1-k)\bm{a}'_{2}\\
    \bm{b}^{k}_{1} &= k\bm{b}_{1} + (1-k)\bm{b}'_{1}\\
    \bm{b}^{k}_{2} &= k\bm{b}_{2} + (1-k)\bm{b}'_{2}
  \end{align}
  ここで, $k$は$0 \leq k \leq 1$となる連続パラメータ, $\{\bm{a}, \bm{b}\}'_{\{1, 2\}}$は$\bm{\theta}_{i+1}$のときのワイヤの始点・終止点である.
  本来は$\bm{\theta}_{i}$から$\bm{\theta}_{i+1}$までの軌道上でそれぞれ$\{\bm{a}, \bm{b}\}^{k}_{\{1, 2\}}$を計算するべきであるが, ここでは計算量の観点から$\{\bm{a}, \bm{b}\}_{\{1, 2\}}$と$\{\bm{a}, \bm{b}\}'_{\{1, 2\}}$の間の線形補間として計算している.
  このとき, $k$を分割して, それぞれの場面において$d^{*} < \epsilon$となるかどうかを判定することで, ワイヤ交差を検知することができる.
  しかし, この場合は$\epsilon$の設定が非常に難しく, 適切に設定しないとワイヤ交差を検知できない可能性がある.
  また, 同様に$k$の分割数の設定も難しく, 粗すぎるとワイヤ交差を検知できない可能性があり, 細かく分割し過ぎると計算量が膨大になる.
  よって本研究では$\epsilon$や$k$の分割数の設定を避けるため, $d^{*}$が下に凸であることを利用し, root-finding algorithmの一種であるBrent法\cite{brent2013algorithms}を用いて$d^{*} < \epsilon$となる$k$を探索する.
  なお, 最大イテレーション数(本研究では20とした)を超えて$d^{*} < \epsilon$に収束しない場合は, ワイヤ交差が起きていないと判定する.

  ここまでは2本のワイヤの交差について述べたが, 実際にはたった2本のワイヤ同士だけでなく, 全てのあり得るワイヤの組み合わせについてこれを行う必要がある.
  折返しがある場合は各ワイヤの線分が交差するかどうかも判定する.
  加えて, ワイヤが関節と交差しないように, 関節リンクの線分とワイヤの間の交差に対しても同様に計算を行う.
}%

\subsection{Joint Torque Evaluation} \label{subsec:torque-eval}
\switchlanguage%
{%
  We describe the method for calculating feasible joint torque along the given trajectory.
  To reduce computational complexity, the feasible joint torque is calculated only at the waypoints $\bm{\theta}_{i}$ ($1 \leq i \leq N_{traj}$) rather than along the entire trajectory.
  In tendon-driven systems, the following equation generally holds,
  \begin{align}
    \bm{\tau} &= -\bm{G}^T(\bm{\theta})\bm{f} \label{eq:torque-tension}
  \end{align}
  where $\bm{f}$, $\bm{\tau}$, and $\bm{G}(\bm{\theta})$ represent the muscle tension, joint torque, and muscle Jacobian, respectively.
  The vector $\bm{f}$ is an $M$-dimensional vector, and $\bm{\tau}$ is a $D$-dimensional vector.
  The space of muscle tensions can be expressed as follows.
  \begin{align}
    F := \{\bm{f} \in \mathbb{R}^{M} \mid \bm{f}^{min} \leq \bm{f} \leq \bm{f}^{max}\} \label{eq:tension-space}
  \end{align}
  By transforming $F$ using \equref{eq:torque-tension}, the joint torque space $T$ can be obtained as follows.
  \begin{align}
    T := \{\bm{\tau} \in \mathbb{R}^{N} \mid \exists\bm{f} \in F, \bm{\tau} = -G^T\bm{f}\} \label{eq:torque-space}
  \end{align}
  To maintain a given joint angle $\bm{\theta}_{i}$, it is necessary to generate joint torques in all directions at that state.
  This requires that a hypersphere $B$ with a radius $R_B$ ($R_B > 0$) and centered at the origin is inscribed within the joint torque space $T$.
  Conversely, if the origin lies outside of $T$, it indicates that sufficient joint torque cannot be generated.

  We now describe how to determine whether the origin lies within $T$, and if so, how to calculate the radius $R_B$ of the inscribed hypersphere.
  First, we extract all the vertices of the hypercube $F$ formed by $\bm{f}$ and project them into the joint torque space $T$.
  We then compute the convex hull $C$ of all the projected vertices.
  To check whether the origin is inside $C$, we compute the signed distance $d_C$ from the origin to all the hyperplanes of $C$.
  If the origin is inside $C$, all $d_C$ values are negative.
  Therefore, if the maximum value of $d_C$ is zero or greater, the origin is not inside $C$.
  If the origin is inside $C$, the minimum absolute value of $d_C$ represents the radius $R_B$ of the inscribed hypersphere.
  Although the effect of gravity is not directly considered in this study, if gravity were to be considered, the origin of the hypersphere can be shifted to the gravity compensation torque value.
  % Furthermore, since only quasi-static situations are considered, the dynamics of the wires, including slack and viscous friction that become problematic during dynamic movements, are not addressed.
}%
{%
  ここでは, 与えられた軌道上における発揮可能関節トルクを計算する方法について述べる.
  計算量削減のため, 軌道上全てではなく, 経由点である$\bm{\theta}_{i}$ ($1 \leq i \leq N_{traj}$)における発揮可能関節トルクを計算する.
  まず, 腱駆動構造においては以下の式が一般に成り立つ.
  \begin{align}
    \bm{\tau} &= -\bm{G}^T(\bm{\theta})\bm{f} \label{eq:torque-tension}
  \end{align}
  ここで, $\bm{f}$は筋張力, $\bm{\tau}$は関節トルク, $\bm{G}(\bm{\theta})$は筋長ヤコビアンを表す.
  $\bm{f}$は$M$次元のベクトル, $\bm{\tau}$は$D$次元のベクトル($D$は関節の自由度数を表す)である.
  筋張力の空間は以下のように表現できる.
  \begin{align}
    F := \{\bm{f} \in \mathbb{R}^{M} \mid \bm{f}^{min} \leq \bm{f} \leq \bm{f}^{max}\} \label{eq:tension-space}
  \end{align}
  $F$を\equref{eq:torque-tension}によって変換することで, 関節トルク空間$T$が以下のように求まる.
  \begin{align}
    T := \{\bm{\tau} \in \mathbb{R}^{N} \mid \exists\bm{f} \in F, \bm{\tau} = -G^T\bm{f}\} \label{eq:torque-space}
  \end{align}
  ある関節角度$\bm{\theta}_{i}$を維持するためには, その状態で全ての関節方向に対して関節トルクが発揮可能である必要がある.
  つまり, 原点を中心とする半径$R_B$ ($R_B>0$)の超球$B$が$T$に内接する必要がある.
  逆に, 原点が$T$の外部に存在する場合, 関節トルクを確保できない.

  ここで, 原点が$T$の内部に存在するかどうか, また存在する場合の内接超球の半径$R_B$を計算する方法について述べる.
  まず, $\bm{f}$が張る超立方体$F$の全頂点を取り出し, それを$\bm{\tau}$の空間$T$へと射影する.
  そして, その射影された全頂点のconvex hull $C$を計算する.
  この$C$内に原点が含まれているかを判定する.
  $C$の全超平面について, 原点からの符号付き距離$d_C$を求める.
  原点が$C$の内部に存在する場合$d_C$は全て負になるため, $d_C$の最大値が0以上であれば原点は$C$の内部に含まれない.
  原点が$C$の内部に含まれる場合, $d_C$の絶対値の最小値が内接超球の半径$R_B$となる.

  なお, 本研究では重力の影響は直接考慮していないが, 考慮する場合は超球の原点を重力補償トルクとすれば良い.
  また, 準静的な状況のみを考えているため, 動的な動きで問題となるワイヤの緩みや粘性摩擦を含むワイヤのダイナミクスは扱っていない.
}%

\subsection{Multi-Objective Black-Box Optimization} \label{subsec:optimization}
\switchlanguage%
{%
  The wire arrangement is optimized using multi-objective black-box optimization.
  For the current wire arrangement, the evaluation functions are set based on the number of wire crossings along the trajectory and the feasible joint torque.
  The optimization aims to minimize the former and maximize the latter.
  The evaluation functions are defined as follows,
  \begin{align}
    E_{cross} &= \sum{h_{cross}(d^{*} < \epsilon)} \label{eq:cross-eval} \\
    E_{torque} &= \prod{h_{torque}(C)} \label{eq:torque-eval}
  \end{align}
  where $h_{cross}$ is a function that returns 1 if a wire crossing occurs and 0 otherwise, and $E_{cross}$ represents the total number of wire crossings along the trajectory.
  Also, $h_{torque}$ returns $R_{min}$ ($1.0\times10^{-3}$ in this study) if the origin is not inside $C$, and $R_B$ if the origin is inside $C$.
  $E_{torque}$ represents the product of the radii of the inscribed hyperspheres of feasible joint torque at all waypoints along the trajectory.
  Using the product ensures that $R_{B}$ is uniformly maximized across all postures.
  While it is also possible to use the summation, this approach may result in optimization where $R_{B}$ is maximized for certain postures but minimized for others.
  Using these two evaluation functions, we optimize the wire arrangement with NSGA-II, a multi-objective optimization algorithm implemented in the black-box optimization library Optuna \cite{akiba2019optuna}.
  NSGA-II is chosen for its ability to handle multi-objective optimization and its relatively large recommended sample size.
  For all experiments, the number of iterations is set to 30,000.
  The optimization took approximately two hours for $N=2$, $M=3$, and $D=2$, and seven hours for $N=3$, $M=6$, and $D=3$.
}%
{%
  ここでは, ワイヤ配置の最適化を多目的ブラックボックス最適化により行う.
  現在のワイヤ配置に対して, 軌道上におけるワイヤ交差の回数と発揮可能関節トルクを評価関数として設定し, 前者を最小化, 後者を最大化するようにワイヤ配置を探索する.
  よって, 以下のように評価関数を定義する.
  \begin{align}
    E_{cross} &= \sum{h_{cross}(d^{*} < \epsilon)} \label{eq:cross-eval} \\
    E_{torque} &= \prod{h_{torque}(C)} \label{eq:torque-eval}
  \end{align}
  ここで, $h_{cross}$はワイヤ交差が起きた場合に1, そうでない場合に0を返す関数であり, $E_{cross}$は軌道上の全てのワイヤ交差の合計値を表している.
  また, $h_{torque}$は原点が$C$の内部に含まれない場合に$R_{min}$ (本研究では$R_{min}=1.0\times10^{-3}$とした), 含まれる場合に$R_B$を返す関数である.
  $E_{torque}$は軌道上の全ての経由点における発揮可能関節トルクの内接超球の半径の積を表している.
  積を用いることで, 全ての姿勢で$R_{B}$を均等に大きくするように最適化される.
  和を用いることも可能だが, その場合は一部の姿勢のみで$R_{B}$を大きく, 他の姿勢で$R_{B}$を小さくするような最適化が起きてしまう.
  これら2つの評価関数をもとに, ブラックボックス最適化のライブラリであるoptuna \cite{akiba2019optuna}に実装された多目的最適化手法であるNSGA-IIを用いてワイヤ配置の最適化を行う.
  このNSGA-IIは, 多目的最適化ができることと, 推奨されるサンプリング数が比較的大きいことから選択した.
  全ての実験についてイテレーション回数は30000とし, 最適化は$N=2, M=3, D=2$で2時間, $N=3, M=6, D=3$で7時間程度かかる.
}%

\begin{figure*}[t]
  \centering
  \includegraphics[width=1.8\columnwidth]{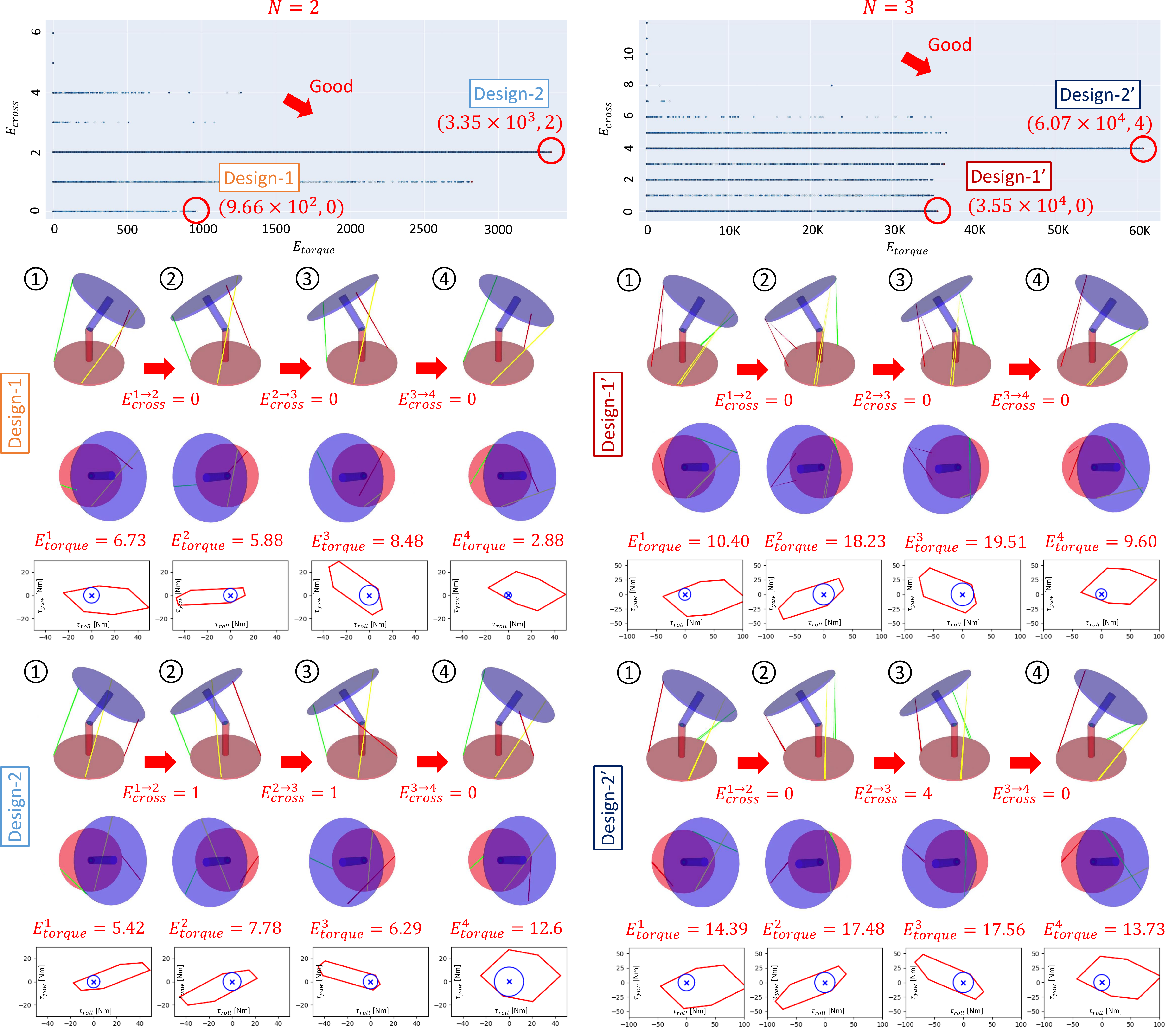}
  \vspace{-1.0ex}
  \caption{The sampling results, design solutions, and torque evaluation for simulation experiments on a 2-DOF joint at $N=\{2, 3\}$ and $M=3$.}
  \vspace{-3.0ex}
  \label{figure:exp-12-m3}
\end{figure*}

\begin{figure*}[t]
  \centering
  \includegraphics[width=1.8\columnwidth]{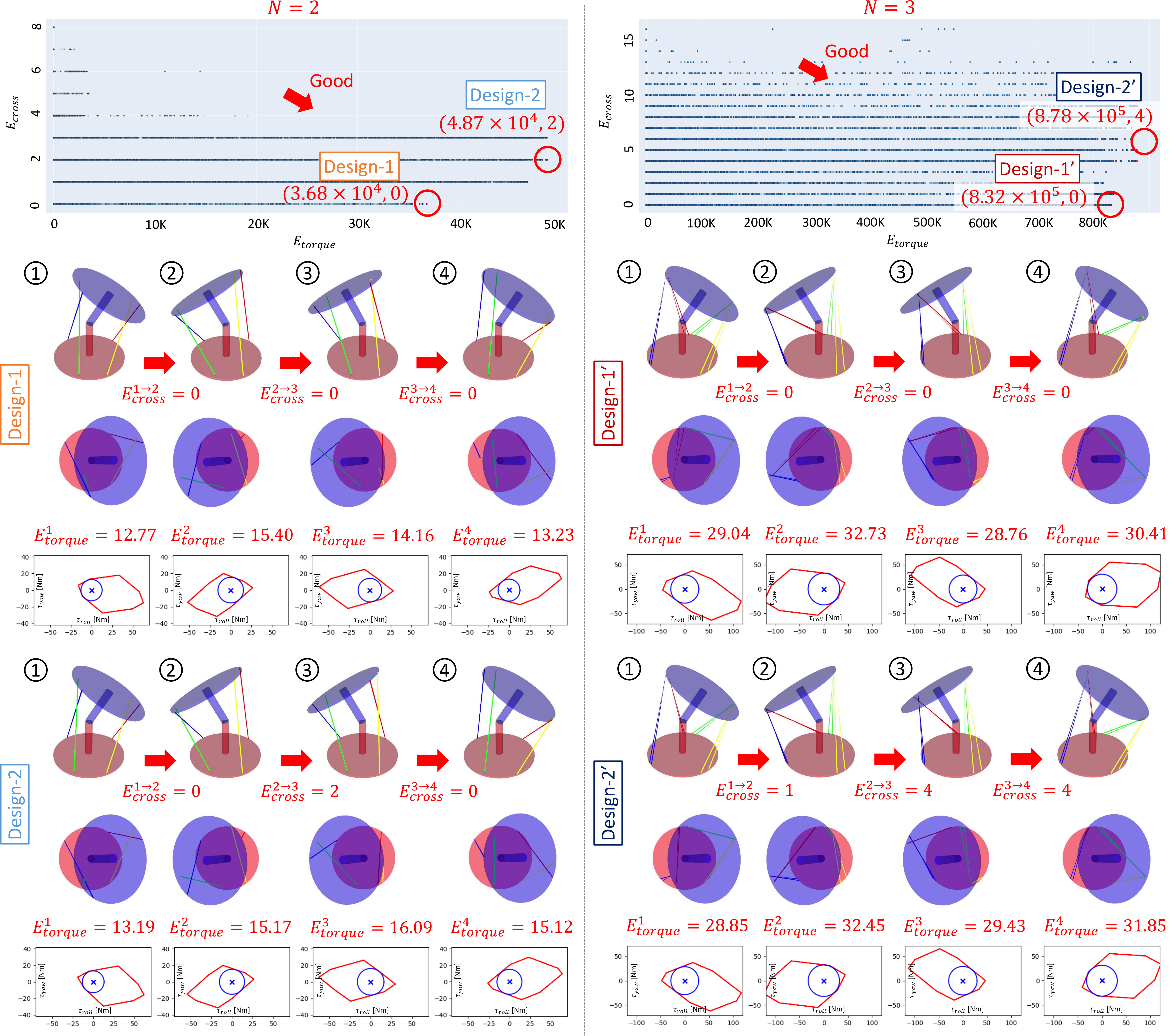}
  \vspace{-1.0ex}
  \caption{The sampling results, design solutions, and torque evaluation for simulation experiments on a 2-DOF joint at $N=\{2, 3\}$ and $M=4$.}
  \vspace{-3.0ex}
  \label{figure:exp-12-m4}
\end{figure*}

\section{Simulation Experiments} \label{sec:experiment}

\subsection{Experimental Setup} \label{subsec:exp-setup}
\switchlanguage%
{%
  We describe the experimental setup of this study.
  The 3D link structure was set with parameters $R = 0.2$ [m] and $L = 0.2$ [m], and the muscle tension limits were set as $f^{min} = 1.0$ [N] and $f^{max} = 200$ [N].
  The number of joints was either 2-DOF or 3-DOF, where the 2-DOF joints included roll and yaw joints, and the 3-DOF joints included roll, pitch, and yaw joints.
  The number of wires was set to $M = \{3, 4\}$ for the 2-DOF joints and $M = \{4, 5, 6\}$ for the 3-DOF joints, and the number of relay points was set to $N = \{2, 3\}$.

  The joint angle trajectories were set as follows: for the 2-DOF joints, $\{(30, 30)$, $(-30, 30)$, $(-30, -30)$, $(30, -30)\}$ [deg]; and for the 3-DOF joints, $\{(20, 20, 30)$, $(20, 20, -30)$, $(-20, 20, -30)$, $(-20, 20, 30)$, $(-20, -20, 30)$, $(-20, -20, -30)$, $(20, -20, -30)$, $(20, -20, 30)\}$ [deg].
  In this study, the trajectory covers a wide range of motion evenly, but task-specific trajectories could also be used.

  Here, we explain how to interpret the results using \figref{figure:exp-12-m3}.
  This figure shows the experimental results for the 2-DOF joints with 3 wires and 2 or 3 relay points.
  The topmost graph displays the sampling results from the optimization process, with the $x$-axis representing $E_{torque}$ and the $y$-axis representing $E_{cross}$.
  Lighter colored points show initial solutions, while darker colored points show final solutions.
  The solutions located in the lower right part of the graph are the better solutions.
  Among the Pareto solutions, the design with $E_{cross} = 0$ (i.e., no wire crossings) is labeled as Design-1, and the design with $E_{cross} > 0$ and the largest $E_{torque}$ is labeled as Design-2.
  Similarly, when $N = 3$, the designs are labeled as Design-1' and Design-2', respectively.
  For these designs, the lower part of the figure shows the motion of the robots and the graphs of the feasible joint torque regions.
  \ctext{1}--\ctext{4} show the robots at the joint angles $\{\bm{\theta}_1, \cdots, \bm{\theta}_{4}\}$ from side and top views.
  $E^{{i}\rightarrow{i+1}}_{cross}$ indicates the number of wire crossings along the trajectory from $\bm{\theta}_{i}$ to $\bm{\theta}_{i+1}$.
  Note that what may appear as a crossing between wires or between a wire and a link from one view often does not represent an actual crossing when viewed from another perspective.
  The feasible joint torque region for each joint angle is shown in the graph, with the $x$-axis representing the torque in the roll direction and the $y$-axis representing the torque in the yaw direction.
  The convex hull $C$ in \secref{subsec:torque-eval} is shown by the red line, and the inscribed hypersphere $B$ is shown by the blue line.
  $E^{i}_{torque}$ indicates the radius $R_B$ of the inscribed hypersphere of the feasible joint torque at $\bm{\theta}_{i}$.
  For the 3-DOF joints, \ctext{1}--\ctext{8} show the robots at the joint angles $\{\bm{\theta}_1, \cdots, \bm{\theta}_{8}\}$, and the feasible joint torque region is plotted with the $x$-axis as the roll direction and the $y$-axis as the yaw direction, with the pitch torque not shown.
  In other words, the 3D convex hull $C$ and the inscribed hypersphere $B$ are projected onto a 2D plane.
}%
{%
  本研究の実験設定について述べる.
  まず, 3次元リンク構造は$R=0.2$, $L=0.2$と設定し, 筋張力については$f^{min}=1.0$ [N], $f^{max}=200$ [N]とした.
  関節数は2自由度または3自由度であり, 2自由度の場合はrollとyaw, 3自由度の場合はroll, pitch, and yawとしている.
  また, ワイヤの本数は2自由度関節について$M=\{3, 4\}$, 3自由度関節について$M=\{4, 5, 6\}$とし, 経由点数は$N=\{2, 3\}$として実験を行っている.
  与える関節角度軌道であるが, 2自由度関節の場合は$\{(30, 30)$, $(-30, 30)$, $(-30, -30)$, $(30, -30)\}$ [deg], 3自由度関節の場合は$\{(20, 20, 30)$, $(20, 20, -30)$, $(-20, 20, -30)$, $(-20, 20, 30)$, $(-20, -20, 30)$, $(-20, -20, -30)$, $(20, -20, -30)$, $(20, -20, 30)\}$ [deg]とした.
  本実験では広い範囲をまんべんなく動かす軌道を取っているが, よりタスクに特化した軌道を与えても良い.

  ここで, \figref{figure:exp-12-m3}を使い図の見方について説明する.
  これは2自由度関節において, ワイヤ数を3, 経由点数を2または3とした場合の実験結果である.
  まず, 一番上の図は最適化におけるサンプリング結果であり, $x$軸を$E_{torque}$, $y$軸を$E_{cross}$としてプロットされている.
  イテレーション数が多過ぎて分かりにくいが, 各サンプリング点は, 薄いほど初期の解, 濃いほど最終的な解を示している.
  これは, グラフの右下にある解ほど良い解である.
  パレート解のうち, $E_{cross}=0$, つまりワイヤが一切交差しない設計解をDesign-1, $E_{cross}>0$の中で最も$E_{torque}$が大きい設計解をDesign-2としている.
  なお, $N=3$の場合はそれぞれDesign-1', Design-2'とする.
  これらの解について, 下の段に設計解の動作と発揮可能関節トルク領域のグラフを示している.
  \ctext{1}--\ctext{4}はそれぞれ$\{\bm{\theta}_1, \cdots, \bm{\theta}_{4}\}$における関節角度の状態を関節の横と上から見た図である.
  $E^{{i}\rightarrow{i+1}}_{cross}$は$\bm{\theta}_{i}$から$\bm{\theta}_{i+1}$までの軌道上でのワイヤ交差の回数を示している.
  なお, 一方向から見てワイヤ同士やワイヤとリンクが交差しているように見えても, 別方向から見ると交差していない場合がほとんどである点に注意いただきたい.
  また, 各関節角度における発揮可能関節トルクの領域がグラフで示されており, $x$軸をroll方向, $y$軸をyaw方向のトルクとして, \secref{subsec:torque-eval}における凸包$C$を赤線で, 内接超球$B$を青線で示している.
  $E^{i}_{torque}$は$\bm{\theta}_{i}$における発揮可能関節トルクの内接超球の半径$R_B$を示している.
  なお, 3自由度関節の場合は\ctext{1}--\ctext{8}の関節角度が示されており, 発揮可能関節トルク領域は$x$軸をroll方向, $y$軸をyaw方向としてpitch方向のトルクは示していない.
  つまり, 3次元の凸包$C$と内接超球$B$を2次元平面に射影した形となっている.
}%

\subsection{2-DOF Joint Experiments} \label{subsec:exp-12}
\switchlanguage%
{%
  \figref{figure:exp-12-m3} shows the results for $N=\{2, 3\}$ and $M=3$, while \figref{figure:exp-12-m4} shows the results for $N=\{2, 3\}$ and $M=4$.
  From the sampling results, we can observe several Pareto solutions for both cases.
  Each of these includes solutions with no wire crossings, indicated by $E_{cross}=0$.
  Additionally, we can see that allowing wire crossings leads to design solutions with larger feasible joint torque.

  First, we consider the results for $M=3$.
  The wire arrangement obtained is asymmetric and non-trivial.
  In Design-1, the feasible joint torque space becomes quite small at certain joint angles, while this is improved in Design-2.
  When $N$ is increased to 3, for example, Design-1' consists of two wires that simply fold back from Design-1, while the third wire has start and end positions that differ somewhat.
  Design-2' is mostly a design where the wires from Design-2 are simply folded back.
  Fundamentally, increasing $N$ from 2 to 3 should double the tension of each wire by folding it back, and thus, the radius of the inscribed hypersphere should also double.
  Therefore, based on the definition in \equref{eq:torque-eval}, a design solution with $E_{torque}$ increased by $2^4$, or 16 times, is possible.
  Indeed, comparing Design-2 and Design-2', we see that $E_{torque}$ is approximately 16 times larger, as the design is essentially a folded version of the original.
  On the other hand, Design-1' exhibits an even greater increase in $E_{torque}$ compared to Design-1, demonstrating the effectiveness of the increased design variation afforded by the foldback.

  Next, we consider the results for $M=4$.
  Here, with four wires allocated to two joints, there is relatively more flexibility in wire placement, allowing for larger inscribed hypersphere radii at all joint angles.
  Additionally, the four-wire setup enables symmetrical designs, resulting in all Pareto solutions being symmetric.
  Because of the symmetric design, there is less interference between wires, and as a result, the $E_{torque}$ values for Design-1 and Design-2 are relatively close.
  When $N=3$, the start and end positions of each wire are not significantly different, resulting in a configuration that appears to fold the wires back.
  Indeed, by increasing $N$ from 2 to 3, we observe that $E_{torque}$ increases by approximately 16 times.
}%
{%
  \figref{figure:exp-12-m3}には, $N=\{2, 3\}, M=3$の際の実験結果を, \figref{figure:exp-12-m4}には$N=\{2, 3\}, M=4$の際の実験結果を示す.
  まず, それぞれのサンプリング結果からいくつかのパレート解が出ていることがわかる.
  それらには, どれもワイヤ交差を防いだ$E_{cross}=0$の解が存在する.
  またワイヤ交差を許せば, より大きな発揮可能関節トルクを持つ設計解が得られることもわかる.

  $M=3$の結果について考察する.
  非対称で非自明なワイヤ配置が得られており興味深い.
  Design-1では一部の関節角度で発揮可能関節トルク空間がかなり小さくなってしまっているが, Design-2ではそれが改善されている.
  $N=3$にすることで, 例えばDesing-1'は3本の筋肉のうち2本の筋はDesign-1をそのまま折り返しただけの形, うち1本は始点と終点の位置がある程度異なる設計となっている.
  Design-2'はほとんどDesign-2のワイヤをそのまま折り返しただけの設計となっている.
  基本的に, $N=2$から$N=3$に増やすことで, ワイヤをそのまま折り返せば各ワイヤ張力が2倍, つまり内接超球の半径も2倍になるはずである.
  そのため, \equref{eq:torque-eval}の定義から$2^4$, つまり16倍の$E_{torque}$を持つ設計解が可能である.
  実際に比べてみると, Design-2とDesign-2'はそのまま折り返したような設計のため, $E_{torque}$は16倍程度になっている.
  一方で, Design-1'はDesign-1に比べると16倍以上の$E_{torque}$を持つ設計解が出ており, 折返しによる設計バリエーションの増加の効果が発揮されている.

  次に$M=4$の結果について考察する.
  こちらは2関節に4本のワイヤを配置するため, ワイヤ本数に比較的余裕があり, どの関節角度においても大きな内接球半径が確保できている.
  また, ワイヤが4本のため対称な設計が可能であり, パレート解はどれも対称な設計となっている.
  対称な設計の結果, ワイヤ同士が干渉しにくいため, 比較的Design-1とDesign-2の$E_{torque}$が近い値となっている.
  $N=3$では, 各ワイヤの始点と終点の位置はそこまで大きく異ならず, そのままワイヤを折り返したような形になっている.
  実際に, $N=2$から$N=3$にすることで, $E_{torque}$が約16倍程度になっている.
}%

\begin{figure*}[t]
  \centering
  \includegraphics[width=1.75\columnwidth]{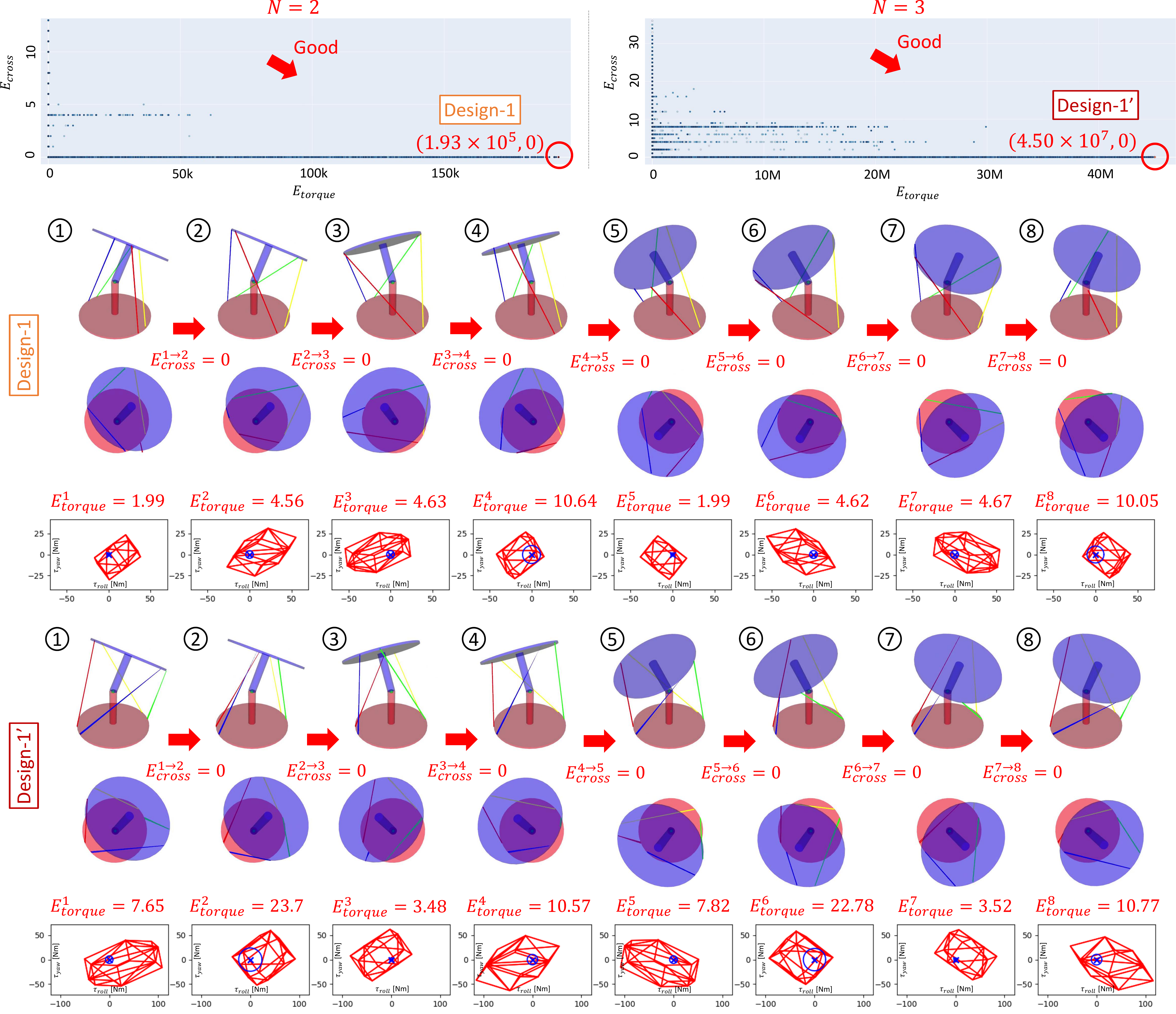}
  \vspace{-1.0ex}
  \caption{The sampling results, design solutions, and torque evaluation for simulation experiments on a 3-DOF joint at $N=\{2, 3\}$ and $M=4$.}
  \vspace{-1.0ex}
  \label{figure:exp-34-m4}
\end{figure*}

\subsection{3-DOF Joint Experiments} \label{subsec:exp-34}
\begin{figure*}[t]
  \centering
  \includegraphics[width=1.75\columnwidth]{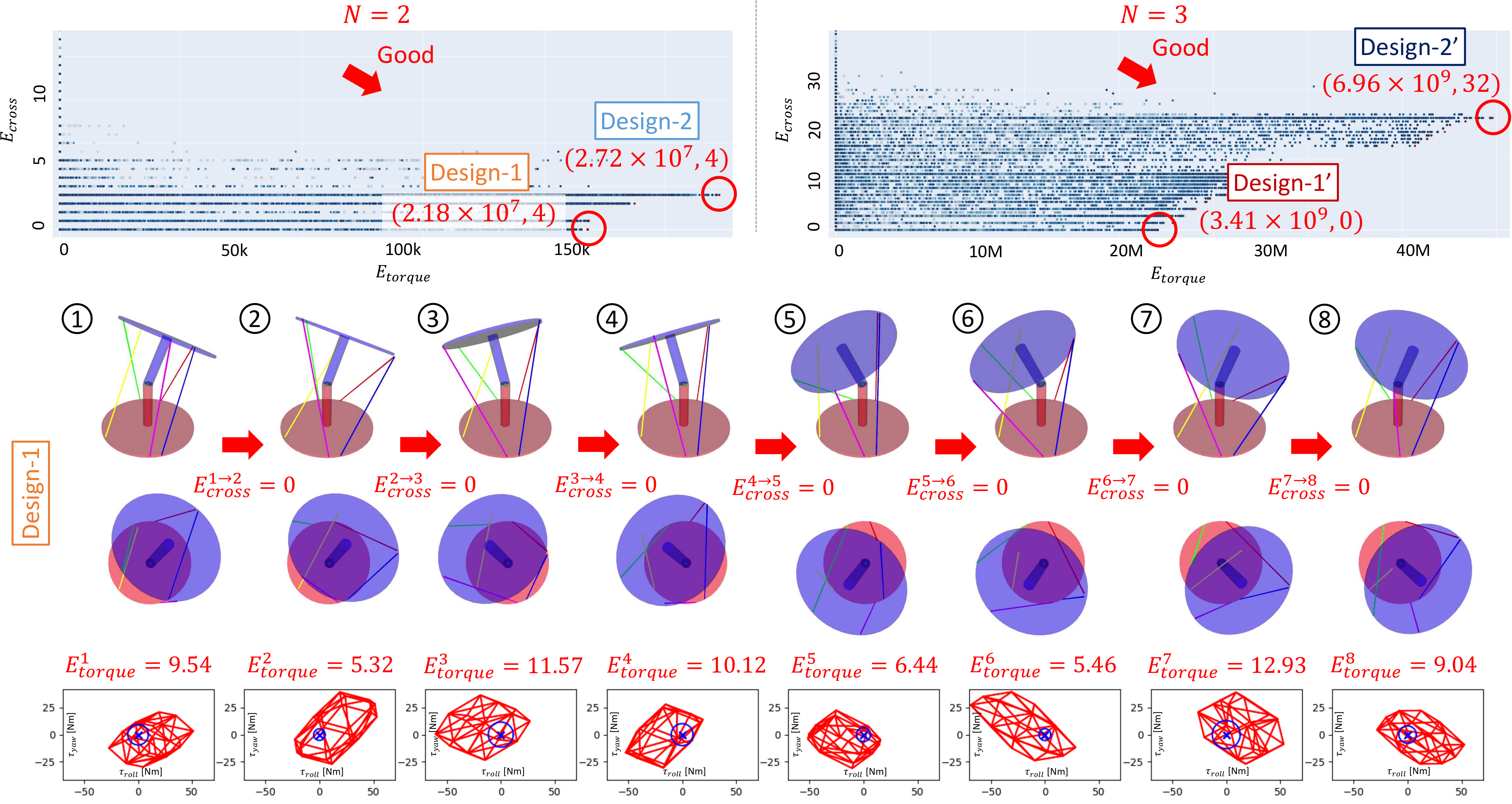}
  \vspace{-1.0ex}
  \caption{The sampling results, design solutions, and torque evaluation for simulation experiments on a 3-DOF joint at $N=\{2, 3\}$ and $M=5$.}
  \vspace{-3.0ex}
  \label{figure:exp-34-m5}
\end{figure*}

\begin{figure*}[t]
  \centering
  \includegraphics[width=1.75\columnwidth]{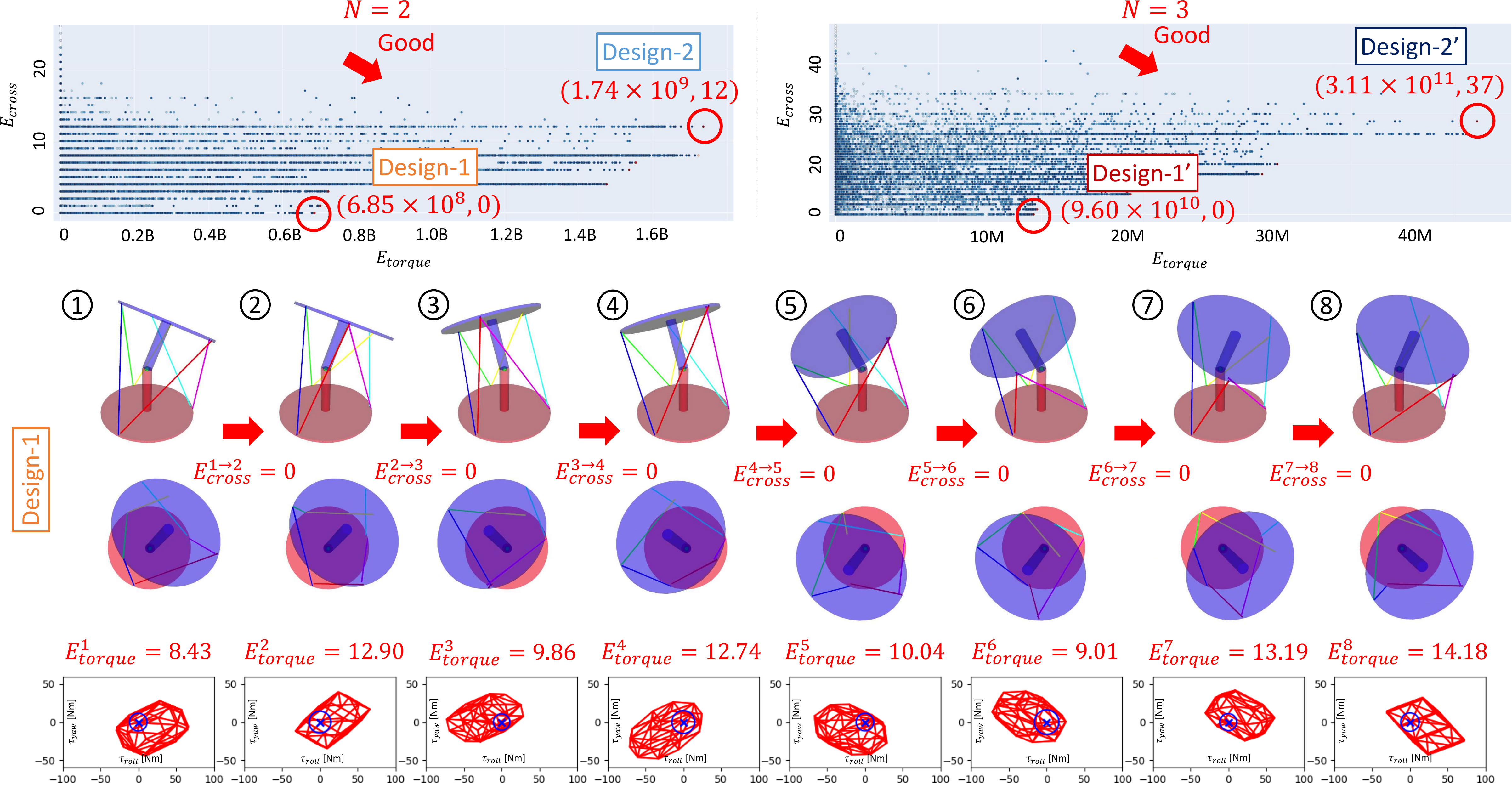}
  \vspace{-1.0ex}
  \caption{The sampling results, design solutions, and torque evaluation for simulation experiments on a 3-DOF joint at $N=\{2, 3\}$ and $M=6$.}
  \vspace{-4.0ex}
  \label{figure:exp-34-m6}
\end{figure*}

\switchlanguage%
{%
  \figref{figure:exp-34-m4} shows the results for $N=\{2, 3\}$ and $M=4$, \figref{figure:exp-34-m5} shows the results for $N=\{2, 3\}$ and $M=5$, and \figref{figure:exp-34-m6} shows the results for $N=\{2, 3\}$ and $M=6$.
  For $M=4$, only the solutions Design-1 and Design-1' are shown, and for $M=\{5, 6\}$, only the Design-1 solutions are shown.
  From the Pareto solutions, we can see that symmetric designs are observed for $M=4$ and $M=6$, whereas no symmetry is observed in the designs for $M=5$.

  For $M=4$, there is only one Pareto solution for both $N=2$ and $N=3$, and we can see that allowing wire crossings does not lead to further improvement in the solution.
  When $N=3$, the start and end points of each wire are so closely aligned that it is difficult to notice any difference in the design solutions.
  By increasing $N$ from 2 to 3, according to the definition in \equref{eq:torque-eval}, it is possible to obtain a design solution with $E_{torque}$ that is $2^8$, or 256 times larger.
  In fact, Design-1' has an $E_{torque}$ approximately 256 times larger than Design-1, indicating that the effect of allowing wire foldback yields minimal benefit.

  For $M=\{5, 6\}$, we obtain highly complex design solutions that are beyond what a human could easily conceive.
  By increasing $N$ from 2 to 3, $E_{torque}$ increases by approximately 256 times, resulting in a design where the wires are mostly folded back without significant changes to their original configuration.
}%
{%
  \figref{figure:exp-34-m4}には$N=\{2, 3\}, M=4$の際の実験結果を, \figref{figure:exp-34-m5}には$N=\{2, 3\}, M=5$の際の実験結果を, \figref{figure:exp-34-m6}には$N=\{2, 3\}, M=6$の際の実験結果を示す.
  なお, ここでは$M=4$についてはDesign-1とDesing-1', $M=\{5, 6\}$についてはDesign-1の解のみを示している.
  パレート解から, $M=4$と$M=6$では設計解に対称性があるが, $M=5$では対称性が見られないことがわかる.

  $M=4$のとき, $N=2$でも$N=3$でもパレート解は一つであり, ワイヤ交差を許してもそれ以上解が改善されないことがわかる.
  $N=3$では設計解を見てもわからないほど, 各ワイヤの始点と終点が一致している.
  $N=2$から$N=3$に増やすことで, \equref{eq:torque-eval}の定義から本実験では$2^8$, つまり256倍の$E_{torque}$を持つ設計解が可能である.
  実際Design-1'はDesign-1よりも$E_{torque}$が約256倍程度であり, ワイヤ折り返しを許すことによる効果はほとんど得られないことがわかる.

  $M=\{5, 6\}$のとき, 人間では考えつかないような複雑な設計解を得ることができている.
  $N=2$から$N=3$に増やすことで$E_{torque}$は256倍程度となっており, ほとんどそのままワイヤを折り返した設計が得られている.
}%

\section{Discussion} \label{sec:discussion}
\switchlanguage%
{%
  Throughout the study, we found that by varying the number of wires and the degrees of freedom of the joints, it is possible to obtain complex and diverse three-dimensional wire arrangements.
  Notably, when the number of wires is odd, we obtain unique design solutions lacking symmetry, which are difficult for humans to conceive.
  In contrast, when the number of wires is even, the solutions tend to be symmetrical and aesthetically pleasing.
  By preventing wire crossings, we can obtain design solutions that are feasible for actual hardware implementation.
  On the other hand, solutions that allow wire crossings often yield better performance.
  This outcome, however, depends on the problem setup, as there are cases where the best performance is achieved without wire crossings, or where performance does not significantly change whether wire crossings are allowed or not.
  Additionally, in the settings of this study, increasing the number of relay points for wires rarely led to performance improvement, and the optimal solutions often involved simply folding the wire back directly.
  While some performance improvements were observed, their impact was small compared to the increased complexity of the design.

  The link structure in this study uses a general shape, without delving into the specifics, because further adjustments are required to link lengths, thicknesses, and even the overall structure itself due to hardware constraints, when designing an actual robot.
  However, We will describe the behavior trends from preliminary experiments when changing the link structure parameters, particularly $R$, $L$, and the joint angle range.
  Increasing $R$ makes wire placement easier, but it narrows the feasible joint angle range.
  Increasing $L$ makes wire placement more difficult and increases the likelihood of wire crossings.
  Regarding the joint angle range, the experiment was conducted within a narrow range of 20 or 30 degrees.
  However, expanding the range further often leads to situations where no solution can be found that maintains joint torque throughout the entire trajectory, especially when the number of wires is small.
  Therefore, in this study, the joint angle range and motion trajectory were chosen to ensure that a reasonable solution can be found.
}%
{%
  全体を通して, ワイヤの本数や関節の自由度数を変えることで複雑で様々な3次元ワイヤ配置が得られることがわかった.
  特にワイヤ本数が奇数のときは対称性のない人間が思いつきづらいユニークな設計解が得られ, 偶数のときは対称性のある綺麗な設計が得られている.
  ワイヤ交差を防ぐことで実際にハードウェアとして実現可能な設計解を得ることができる.
  一方で, ワイヤ交差を許した解の方がより性能の良い解が得られる.
  ただし, それは問題設定次第であり, ワイヤ交差しない解の性能が最も良いことや, ワイヤ交差してもしなくても大きく性能が変化しない場合もある.
  また本研究の設定では, ワイヤの経由点を増やすことが性能向上に有効なことは少なく, 最適解はワイヤをそのまま折り返す解である場合が多い.
  確かに一部性能向上が見られたが, 設計の複雑さに比べて効果は小さい.

  今回扱ったリンク構造であるが, 実際にロボットを設計する際には, さらにリンクの長さや太さ, さらにはそもそもの構造自体もハードウェア制約から変化させ, ワイヤ配置最適化を行う必要がある.
  そのため, リンク構造については一般的な形状を取扱い, ここに深入りはしていない.
  一方で, リンク構造のパラメータ,特に$R$, $L$, および関節角度範囲を変化させた際の挙動の傾向については分かっている範囲で記述する.
  $R$のみを大きくすればワイヤ配置は容易になるが, 動作可能な関節角度範囲は狭まる.
  $L$のみを大きくすればワイヤ配置は難しくなり, ワイヤ交差が発生しやすくなる.
  関節角度範囲については, 今回約30度というのは狭い範囲で実験を行ったが, 特にワイヤ本数が少ないときは, これ以上範囲を広げると全軌道上で関節トルクを保つことのできる解が発見できない場合が多々ある.
  このような解を実験として示しても良いが, あまり面白いものではないため, 今回は全軌道上で関節トルクを保つことのできる解が見つかるような関節角度範囲・動作軌道を選んでいる.
}%

\section{CONCLUSION} \label{sec:conclusion}
\switchlanguage%
{%
  We proposed a method for three-dimensional wire arrangement optimization considering wire crossings using black-box multi-objective optimization.
  Previously, wire arrangement optimization had been applied only to relatively simple systems, such as those in two-dimensional planes or systems with constant moment arms where wire crossings did not need to be considered.
  However, our study removes these constraints, making it applicable to more complex structures.
  We represented the wire arrangement by defining the start, relay, and end points of the wires, and geometrically detected wire crossings along the target trajectory, using their count as an evaluation function in the multi-objective optimization.
  Additionally, we maximized the product of the radii of hyper-spheres inscribed within the joint torque space for each posture, thereby achieving stable movement along the desired trajectory.
  We successfully optimized the wire arrangement under various conditions for a three-dimensional link structure, which directly extends the planar wire arrangement optimization into three dimensions, obtaining a body design capable of achieving the target trajectory stably while preventing wire crossings.
  % It is anticipated that more complex and higher-performance tendon-driven three-dimensional structures will be developed in the future.
}%
{%
  本研究では, ワイヤの交差を考慮した3次元のワイヤ配置最適化をブラックボックス多目的最適化により行う手法を提案した.
  これまでは2次元平面上やモーメントアーム一定, またワイヤ交差を考慮する必要のない比較的シンプルで動きの小さな系に対してのみワイヤ配置最適化が行われていたが, 本研究はそれらの制約を取り払ったより複雑な構造に対して適用可能である.
  ワイヤ配置をワイヤの始点・中継点・終止点で表現し, 目標軌道上におけるワイヤ交差を幾何的に検知し, その回数を多目的最適化の評価関数とした.
  また, 各姿勢において発揮可能な関節トルク空間に内接する超球の半径の積を最大化し, 目的軌道上における安定した動作を実現した.
  平面上のワイヤ配置最適化を直接3次元に拡張した3次元リンク構造に対して, 多様な条件でワイヤ配置を最適化し, ワイヤ交差を防ぎつつ目的の軌道を安定して実現可能な身体設計を獲得することに成功した.
  今後より複雑で高い性能を持つ腱駆動型の3次元構造が多数構築されることが期待される.
}%

{
  %\footnotesize
  %\small
  %\bibliographystyle{junsrt}
  \bibliographystyle{IEEEtran}
  \bibliography{main}
}

\end{document}